\documentclass[10pt,letterpaper]{article}
\usepackage[top=0.85in,left=2.75in,footskip=0.75in,marginparwidth=2in]{geometry}

\usepackage[utf8]{inputenc}
\usepackage[superscript,biblabel]{cite}

\usepackage{hyperref}
\usepackage{amsmath}
\usepackage{multirow}
\usepackage{color}

\usepackage{microtype}
\DisableLigatures[f]{encoding = *, family = * }

\usepackage{changepage}

\usepackage[aboveskip=1pt,labelfont=bf,labelsep=period,singlelinecheck=off]{caption}

\makeatletter
\renewcommand{\@biblabel}[1]{\quad#1.}
\makeatother

\usepackage{lastpage,fancyhdr,graphicx}
\usepackage{epstopdf}
\fancyheadoffset[L]{2.25in}
\fancyfootoffset[L]{2.25in}

\definecolor{Gray}{gray}{.25}

\usepackage{graphicx}

\usepackage{sidecap}

\usepackage{wrapfig}
\usepackage[pscoord]{eso-pic}
\usepackage[fulladjust]{marginnote}
\reversemarginpar

\begin{document}
\vspace*{0.35in}

% title goes here:
\begin{flushleft}
{\Large
\textbf\newline{Novel quantitative indicators of digital ophthalmoscopy image quality}
}
\newline
% authors go here:
\\
Chris v Csefalvay\textsuperscript{1,*}
\\
\bigskip
1 \ Starschema Ltd, Budapest, Hungary
\\
\bigskip
* csefalvayk@starschema.net

\end{flushleft}

\section*{Abstract}
With the advent of smartphone indirect ophthalmoscopy, teleophthalmology – the use of specialist ophthalmology assets at a distance from the patient – has experienced a breakthrough, promising enormous benefits especially for healthcare in distant, inaccessible or opthalmologically underserved areas, where specialists are either unavailable or too few in number. However, accurate teleophthalmology requires high-quality ophthalmoscopic imagery. This paper considers three feature families – statistical metrics, gradient-based metrics and wavelet transform coefficient derived indicators – as possible metrics to identify unsharp or blurry images. By using standard machine learning techniques, the suitability of these features for image quality assessment is confirmed, albeit on a rather small data set. With the increased availability and decreasing cost of digital ophthalmoscopy on one hand and the increased prevalence of diabetic retinopathy worldwide on the other, creating tools that can determine whether an image is likely to be diagnostically suitable can play a significant role in accelerating and streamlining the teleophthalmology process. This paper highlights the need for more research in this area, including the compilation of a diverse database of ophthalmoscopic imagery, annotated with quality markers, to train the Point of Acquisition error detection algorithms of the future.

\section{Introduction} % (fold)
\label{sec:introduction}

\begin{wrapfigure}[15]{l}{75mm}
% the number in [] of wrapfigure is optional and gives the number of text lines that should be wrapped around the text. Adjust according to your figures height
\includegraphics[width=75mm]{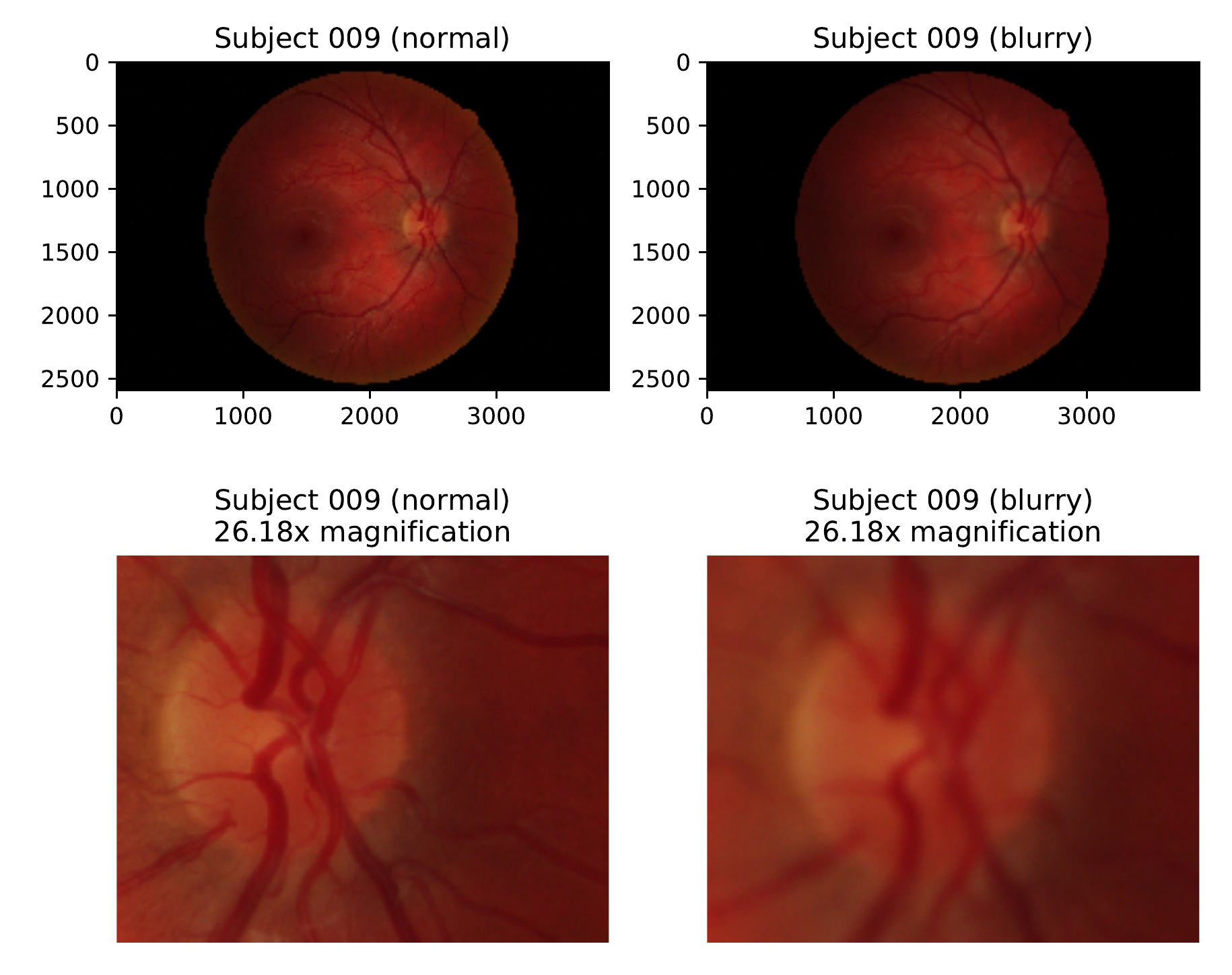}
\captionsetup{labelformat=empty} % makes sure dummy caption is blank
\caption{} % add dummy caption - otherwise \label won't work and figure numbering will not count up
\label{fig1} % use \ref{fig1} to reference to this figure
\end{wrapfigure}

% avoid blank space here 
\marginpar{
\vspace{.7cm} % adjust vertical position relative to text with \vspace{} - note that you can enter negative numbers to move margin caption up
\color{Gray} % this gives caption a grey color to set it apart from text body
\textbf{Figure \ref{fig1}. Comparison of a normal \emph{versus} a blurry ophthalmoscopy image from Köhler et al.\cite{kohler2013automatic}} % note that \ref{fig1} refers to the corresponding wrapfigure
 Note the reduced vascular definition over the optic disc and the obliteration of fine vasculature contrast in the blurry image
}
 % avoid blank space here
With increasing frequency, ophthalmoscopy is performed using digital methods of acquisition, using digital indirect fundus cameras. However, owing to the advent of smartphone digital ophtalmoscopy,\cite{giardini2014smartphone,bastawrous2012smartphone} what was once the preserve of ophthalmoscopic practice in the developed world has become an important tool of the 'democratisation' of eye care and retinal screening, not the least where ophthalmological experts are not available on site or other reasons for telemedicine and image storage prevail.\cite{liesenfeld2000telemedical} The ubiquity of relatively high quality cameras and serious computing power on smartphones has fuelled this trend globally, with incredible ramifications for public health worldwide and the opportunity to screen patients effectively, even in areas where specialised ophthalmological expertise may not be available and where specialised high quality tools for digital ophthalmoscopy are not available, and direct them to adequate treatment. In spite of this, of all the recognised DICOM modalities, fundoscopy imagery (DICOM modality OP) has remained relatively neglected from the standpoint of automated analytics, compared to other more popular modalities, such as magnetic resonance imaging, computed tomography, nuclear medicine and ultrasonography. Considering the potential of teleophthalmology in ophthalmic trauma, ophthalmic infectious disease triage, early recognition of congenital defects and the growing public health threat of diabetic retinopathy,\cite{surendran2014teleophthalmology} the importance of algorithmic analysis of ophthalmic imagery cannot be understated.

From the perspective of machine learning, retinal imagery presents two difficulties. Firstly, the typical ophthalmoscopy image is enormously feature-rich and contextual. Not only are there numerous confounding features, but the clinical meaning of those features is largely contingent on their location. So for instance on a normal, non-pathological image, a slightly hyperpigmented (dark) spot, very slightly smaller (approximately 1.5mm or $5^{\circ}$) than the optic disc (approximately 1.77mm or $5.54^{\circ}$), is positioned roughly halfway between the optic disc and the edge of the fundoscopic image. This is the macula, the area responsible for sharp central colour vision. However, at any other location, the same sort of darker spot might be indicative of localised disease, including e.g. large diffuse haemorrhages occurring in diabetic retinopathy. Simple object recognisers therefore fail at resolving this problem adequately, by struggling to consider that the meaning of a feature is dependent on its milieu (the features surrounding it and its location within the total fundus image). For this reason, most conventional methods of machine learning from images encounter significant obstacles when faced with retinal imagery. 

The \emph{sine qua non} of successful teleophthalmology is the acqusition of high quality retinal imagery. Since teleophthalmology customarily relies on general practitioners, nurses, health visitors and other public health workers without specialist training in ophthalmology and retinal imaging to acquire images that need later be useful for the reviewing ophthalmologist, a Point of Acquisition quality assessment tool is absolutely crucial to ensure the success of teleophthalmology programmes. By far one of the most serious problems is posed by out-of-focus or blurry images – unlike direct ophthalmoscopy, where the clinician can adjust the focus on the ophthalmoscope, digital ophthalmoscopic images are often compromised by blur issues, even those taken using adequate equipment, which can no longer be corrected after acquisition. Image blur may be caused by eye motion during acquisition or, more often, incorrect focus. The problem is even more significant with smartphone ophthalmoscopy, where focus is automatically set by the camera firmware, which was not designed for ophthalmoscopy specifically.

% section introduction (end)

\section{Methodology} % (fold)
\label{sec:methodology}

In order to assess image sharpness, we are making use of three distinct types of metrics. Statistical metrics give overall quantifications of the image, whereas gradient-based metrics are summarised local gradient metrics that are highly locationally sensitive and wavelet transform coefficient based metrics reveal more about the underlying patterns of the image.

\subsection{Statistical metrics} % (fold)
\label{sub:statistical_metrics}

We have employed three statistical metrics from the realm of information theory: overall energy $\Sigma_E$, Shannon entropy $S$ and Atkinson's Entropy Focus Criterion ($\hat{S}$ and $\bar{S}$).\cite{atkinson1997automatic} The latter is more predominantly used in magnetic resonance imaging (MRI), where it serves to identify intra-slice motion artifacts, and to the best of the author's knowledge, this is the first time it has been used to assess blurriness in ophthalmic fundoscopic imagery, albeit its use for iris recognition has been documented previously by Grabowski et al.\cite{grabowski2007focus}

\subsubsection{Energy} % (fold)
\label{ssub:overall_energy}

The overall energy $\Sigma_E$ of an image $\mathbf{I}^{M \times N \times 3}$ is defined as the sum of square pixel intensities. For multi-channel images, both the overall energy of the grayscale image

\begin{equation}
	\Sigma_E(\mathbf{I}^{M \times N}) = \sum_m^M \sum_n^N {\mathbf{I}_{m, n}}^2 
\end{equation}

\noindent can be calculated, as well as the root mean square (RMS) of the channels $C$, so that

\begin{equation}
	\Sigma_E(\mathbf{I}^{M \times N \times 3}) = \sqrt{\frac{1}{C} \ \sum_{ch}^C \left(\sum_m^M \sum_n^N {\mathbf{I}_{m, n, ch}}^2 \right)^2}
\end{equation}

Grayscale conversion is performed using the $Y_{709}$ coefficients (R: $0.2125$, G: $0.7154$, B: $0.0721$), pursuant to ITU-R Recommendation BT.709,\cite{recommendation1990709} using the conversion code in \texttt{skimage.color.rgb2gray}. Conceptually, image energy is the discretised spatial equivalent of the overall spectral energy of a continuous-time signal $\langle x(t), x(t)\rangle$, which is calculated as 

\begin{equation}
	\langle x(t), x(t) \rangle = \int_{-\infty}^{\infty}{\vert x(t) \vert^2} dt
\end{equation}

\noindent This is implemented by the sum of squares representation. To compensate for the size dependence, the mean pixel energy metric

\begin{equation}
	\overline{\Sigma_E} (\mathbf{I}) = \frac{1}{M \ N} \sum_m^M \sum_n^N {\mathbf{I}_{m, n}}^2
\end{equation}

\noindent is used in lieu of overall energy.

% subsubsection overall_energy (end)

\subsubsection{Shannon entropy} % (fold)
\label{ssub:shannon_entropy}

For an image $\mathbf{I}^{M \times N}$, the corresponding Shannon entropy $S(\mathbf{I})$ is defined as the negative $\log_2$ of the number of possible outcomes for the pixels within the image. Shannon entropy can be understood as a metric of the information content in the overall image and is calculated as 

\begin{equation} \label{eq:entropy}
	S(\mathbf{I}^{M \times N}) = - \sum_m^M \sum_n^N \dot{S}(\mathbf{I}_{m, n})
\end{equation}

\noindent where $\dot{S}(x)$ is the element-wise entropy function

\begin{equation*}
	\label{eq:elementwise_entropy}
%	\[
	  \dot{S}(x) = \left.
	  \begin{cases}
	    x \log_2(x), & \text{for } x > 0 \\
	    0, & \text{for } x = 0 \\
	    -\infty, & \text{for } x < 0
	  \end{cases}
	  \right.
%	\]
\end{equation*}

% subsubsection shannon_entropy (end)

\subsubsection{Atkinson's Entropy Focus Criterion (EFC)} % (fold)
\label{ssub:atkinson_s_entropy_focus_criterio}

For an image $\mathbf{I}^{M \times N}$, Atkinson's Entropy Focus Criterion (EFC) $\hat{S}$ is the pixel value entropy, using $\text{e}$ as the basis, normalised by the maximum entropy $S_{max}(\mathbf{I})$, where the entire energy of the image is concentrated in a single pixel.\cite{atkinson1997automatic}. For

\begin{equation}
	S_{max}(\mathbf{I}) = \sqrt{\Sigma_E(\mathbf{I})} = \sqrt{\sum_m^M \sum_n^N {\mathbf{I}_{m, n}}2}
\end{equation}

\noindent the adjusted entropy $\hat{S}$ is defined as

\begin{equation}
	\hat{S} = - \sum_m^M \sum_n^N \frac{\mathbf{I}_{m, n}}{S_{max}(\mathbf{I})} \ ln \left( \frac{\mathbf{I}_{m, n}}{S_{max}(\mathbf{I})} \right)
\end{equation}

% subsubsection atkinson_s_entropy_focus_criterio (end)

\subsubsection{Atkinson's Normalised EFC (NEFC)} % (fold)
\label{ssub:atkinson_s_normalised_efc}

Similarly to Atkinson's EFC (on which see \autoref{ssub:atkinson_s_entropy_focus_criterio}), the normalised EFC $\bar{S}$ is a metric of the relationship between individual pixel values and the maximum image entropy. However, it is robust and invariant in respect of changes in image dimensions by using an adjustment factor. Thus, for an image $\mathbf{I}^{M \times N}$

\begin{equation}
	\bar{S}(\mathbf{I}) = \hat{S}(\mathbf{I}) \ \left( \frac{M \ N}{\sqrt{M \ N}} \text{ln}(M \ N)^{-\frac{1}{2}} \right)
\end{equation}

\noindent where $\hat{S}(\mathbf{I})$ is Atkinson's Entropy Focus Criterion as described in \autoref{ssub:atkinson_s_entropy_focus_criterio}.

% subsubsection atkinson_s_normalised_efc (end)

% subsection statistical_metrics (end)

\subsection{Gradient based metrics} % (fold)
\label{sub:gradient_based_metrics}

In addition to the statistical metrics, we will make use of three primary gradient-based metrics: the sum of thresholded root of squared Sobel kernels (also known as the Tenengrad metric $T(\mathbf{I})$\cite{tenenbaum1970accommodation,schlag1983implementation}), the mean absolute Laplacian (MAL) $\Vert \nabla^2 \Vert (\mathbf{I})$,\cite{laparra2016perceptual,van1989nonlinear} and the log variance of the mean absolute Laplacian $\log \sigma^2_{\Vert \nabla^2 \Vert}$,\cite{pech2000diatom} which I will refer to henceforth as the log Pech-Pacheco metric. In addition, we are using two derived metrics, $ \Vert \nabla^2 \Vert_{\mathcal{V}^\prime}(\mathbf{I})$ and $T_{\mathcal{V}^\prime}(\mathbf{I})$, which are perivasculatity weighted equivalents of $\Vert \nabla^2 \Vert(\mathbf{I})$ and $T(\mathbf{I})$, respectively.

\subsubsection{Tenengrad gradient magnitude} % (fold)
\label{ssub:tenengrad_gradient_magnitude}

Tenengrad is a gradient magnitude metric first described by Tenenbaum\cite{tenenbaum1970accommodation} and later explored by Schlag et al.\cite{schlag1983implementation} and Krotkov\cite{krotkov1988focusing} among others. Tenengrad combines magnitude informations from two Sobel kernels,

\begin{equation}
	\mathbf{S}_{\mathbf{x}} = \begin{bmatrix}
						-1 & 0 & 1 \\
						-2 & 0 & 2 \\
						-1 & 0 & 1 
				   \end{bmatrix}
\end{equation}

\noindent and 

\begin{equation}
	\mathbf{S}_{\mathbf{y}} = \begin{bmatrix}
						1 & 2 & 1 \\
						0 & 0 & 0 \\
						-1 & -2 & -1 
					\end{bmatrix}
\end{equation}

Then, given a source image $\mathbf{I}^{M \times N}$, the gradient magnitude at $\mathbf{I}_{p, q}$, denoted $\mathbf{G}(p, q)$, is calculated as the sum of squared convolutions, i.e. 

\begin{equation}
	\mathbf{G}(p, q) = \sqrt{\left[ \mathbf{G}_x(p, q)\right]^2 + \left[ \mathbf{G}_y(p, q) \right]^2}	
\end{equation}

\noindent where $\mathbf{G}_x(p, q)$ is the convolution of the source image $\mathbf{I}$ with $\mathbf{S_x}$ and $\mathbf{G}_y(p, q)$ is the convolution of the source image $\mathbf{I}$ with $\mathbf{S_y}$, respectively, at the point $\mathbf{I}_{p, q}$. For a threshold value $\tau$, 

\begin{equation}
	\mathbf{G}[\tau](p, q) = \begin{cases} \mathbf{G}(p, q) & \mathbf{G}(p, q) \geq \tau \\ 0 & \mathbf{G}(p, q) < \tau \end{cases} 
\end{equation}

\noindent and the thresholded Tenengrad measure $T_{\tau}(\mathbf{I})$ is then defined as 

\begin{equation}
	T_{\tau}(\mathbf{I}) = \sum_m^M \sum_n^N \left[ \mathbf{G}[\tau](p, q) \right]^2
\end{equation}

\noindent while the unthresholded Tenengrad measure $T(\mathbf{I})$ corresponds to 

\begin{equation}
	T(\mathbf{I}) = \sum_m^M \sum_n^N \left[ \mathbf{G}(m, n) \right]^2
\end{equation}

% subsubsection tenengrad_gradient_magnitude (end)

\subsubsection{Mean of the absolute Laplacian ($\Vert\nabla^2\Vert$) and energy of the Laplacian $E_{\nabla^2}$} % (fold)
\label{ssub:mean_absolute_laplacian}

For an image $\mathbf{I}^{M \times N}$, the Laplacian $\nabla^2(\mathbf{I})$ is defined as

\begin{equation}
	 \nabla^2(\mathbf{I}) = \frac{\partial^2 \mathbf{I}}{\partial x^2} + \frac{\partial^2 \mathbf{I}}{\partial y^2}
\end{equation}

\noindent which in a discrete space can be approximated by the convolution of $\mathbf{I}$ with the kernel 

\begin{equation}\label{eq:nabla_matrix}
    \mathbf{K}_{\nabla^2} = \frac{1}{6} \ \begin{bmatrix}
0 & -1 & 0 \\
-1 & 4 & -1 \\
0 & -1 & 0 
\end{bmatrix}
\end{equation}

\noindent Then, the mean of absolute values of the Laplacian over the entire image space $M \times N$ are

\begin{equation}
	\Vert \nabla^2 \Vert (\mathbf{I}) = \frac{1}{M N} \sum_m^M \sum_n^N \vert \mathbf{K}_{\nabla^2} (m, n) \vert	
\end{equation}

\noindent where $\mathbf{K}_{\nabla^2} (m, n)$ denotes the convolution of $\mathbf{I}_{m, n}$ with the kernel $\mathbf{K}_{\nabla^2}$ (see \autoref{eq:nabla_matrix}). For an image $\mathbf{I}^{M \times N}$, the energy of the Laplacian $E_{\nabla^2}$ is defined as

\begin{equation}\label{eq:energy_of_laplacian}
\sum_m^M \sum_n^N \left({\nabla^2 (\mathbf{I})_{m, n}}\right)^2
\end{equation}

% subsubsection mean_absolute_laplacian (end)

\subsubsection{Logarithmic Pech-Pacheco (Laplacian log variance) metric ($\log \sigma^2_{\Vert \nabla^2 \Vert}$)} % (fold)
\label{ssub:log_variance_of_vert_nabla_2_vert}

Given an image $\mathbf{I}^{M \times N}$, the log variance of the mean of the absolute Laplacian is defined, by reference to $\Vert \nabla^2 \Vert$ (see \autoref{ssub:mean_absolute_laplacian}), as

$$ \log \sigma^2_{\Vert \nabla^2 \Vert} = \log \left[ \sum_m^M \sum_n^N \left( \vert \mathbf{K}_{\nabla^2}(m, n) \vert - \Vert \nabla^2 \Vert (\mathbf{I}) \right)^2 \right] $$

\noindent where $\mathbf{K_{\nabla^2}}$ is the Laplacian kernel described in \autoref{eq:nabla_matrix}.

% subsubsection log_variance_of_vert_nabla_2_vert (end)

\subsubsection{Mean perivascular Tenengrad and absolute Laplacian} % (fold)
\label{ssub:mean_perivascular_tenengrad_and_absolute_laplacian}

The perivascular area – the sharp margin between the relatively homogenous blood vessel and the equally rather homogenous retinal surface – is an outstanding indicator of image definition, as it shows how well an unconditional ground truth (the sharp vascular margin) is reflected in the image. Therefore, perivascular weighting of Tenengrad (see \autoref{ssub:tenengrad_gradient_magnitude}) and absolute Laplacian (see \autoref{ssub:mean_absolute_laplacian}) metrics can give a better image of image sharpness and definition. To facilitate this, a perivascular mask $\mathcal{V}^\prime(\mathbf{I})$ is constructed, which is then used to weight the vascularity-naive $\mathbf{G}[\tau](\mathbf{I})$ and $\Vert \nabla^2 \Vert(\mathbf{I})$ metrics. The results are then re-averaged to arrive at the mean perivascular Tenengrad magnitude $T_{\mathcal{V}^\prime}(\mathbf{I})$ and the mean perivascular absolute Laplacian $ \Vert \nabla^2 \Vert_{\mathcal{V}^\prime}(\mathbf{I})$, respectively.

\vspace{1.5cm}

\begin{figure}[ht] %s state preferences regarding figure placement here

% use to correct figure counter if necessary
%\renewcommand{\thefigure}{2}

\includegraphics[width=\textwidth]{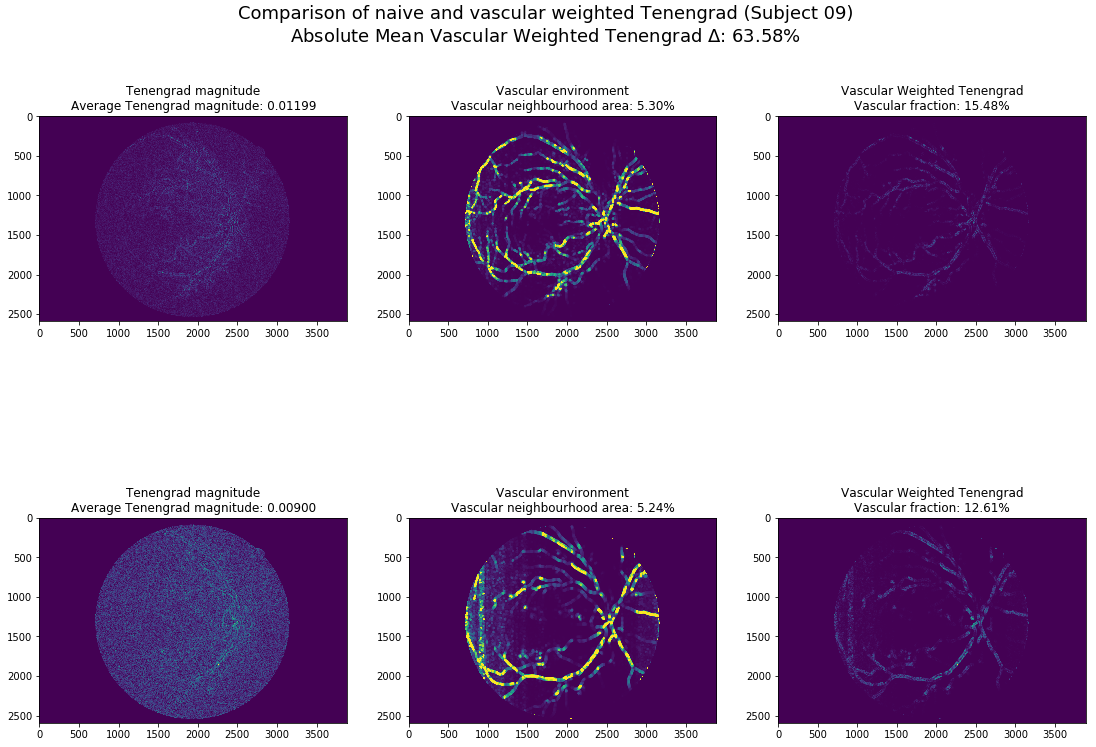}

\caption{\color{Gray} \textbf{Perivascular mask weighting the Tenengrad gradient magnitude (see \autoref{ssub:tenengrad_gradient_magnitude}) of two exposures of the same subject (Subject 09 from Köhler et al.\cite{kohler2013automatic}).} Note that even with a relatively small difference in perivascular mask size, the weighted Tenengrad gradient (the product of the unnormalised Tenengrad magnitude multiplied by the perivascular environment matrix) exhibits a significant (63.58\%) difference between the sharp (top) and blurry (bottom) images. This accentuates the utility of perivascular weighting of gradient-based metrics.}

\label{fig2} % \label works only AFTER \caption within figure environment

\end{figure}

\clearpage

The perivascular mask $\mathcal{V}^\prime$ is computed by reference to a filtered image $\mathcal{F}(\mathbf{I})$, where $\mathcal{F}$ is the multiscale vessel enhancement filter described by Frangi et al. (1998).\cite{frangi1998multiscale} This yields the perivascular mask $\mathcal{V}^\prime$ so that

\begin{equation}
	\mathcal{V}^\prime (\mathbf{I})_{m, n} = \sqrt{{\hat{A}_{m, n}(\mathcal{F}(\mathbf{I}))_{1, 0}}^2 + {\hat{A}_{m, n}(\mathcal{F}(\mathbf{I}))_{1, 1}}^2 + {\hat{A}_{m, n}(\mathcal{F}(\mathbf{I}))_{0, 1}}^2}
\end{equation}

\noindent where $\hat{A}(\mathbf{I})$ is the structure tensor of the image $\mathbf{I}$ at $m, n$ with $\sigma = 1$. This creates a 'vascular neighbourhood' image that can be used to weight gradients in the vascular environment. Since the vascular environment tends to contain sharp edges (the vascular boundary, which in a sharp ophthalmoscopic photograph appears as a steep gradient), weighting the vascular neighbourhood can 'focus' gradient metrics (for an example, see Figure 2).

% subsubsection mean_perivascular_tenengrad_and_absolute_laplacian (end)

% subsection gradient_based_metrics (end)

\subsection{Wavelet coefficients} % (fold)
\label{sub:wavelet_coefficients}

Let $\mathcal{W}_{\psi}(\mathbf{I})$ be the discrete 2D wavelet transform of $\mathbf{I}$ with the wavelet $\psi$, with coefficients $\psi_{1 \cdots n}$. Then, the ordered set $\left\{ \mathcal{W}_{\psi}(\mathbf{I}) \right\}^{M \times N \times 3}$ is the ordered set of the horizontal, vertical and diagonal coefficients of the wavelet transform. Then, for $\left\{ \mathcal{W}_{\psi}(\mathbf{I}) \right\}^{M \times N \times 3}$, let the variance of wavelet coefficients $\sigma^2 \left( \left\{ \mathcal{W}_{\psi}(\mathbf{I}) \right\} \right)$ be defined as the ordered set

\begin{equation}\label{eq:variance_of_wavelets}
\sigma^2 \left( \left\{ \mathcal{W}_{\psi}(\mathbf{I}) \right\} \right) = \left\{\left. \frac{1}{M \ N} \sum_m^M \sum_n^N \left( \mathbf{W}_{m, n} - \overline{\mathbf{W}} \right)^2 \ \right\vert \ \mathbf{W} \in \mathcal{W}_{\psi}(\mathbf{I}) \right\}
\end{equation}

\noindent where

\begin{equation}\label{eq:mean_of_wavelet_coeff}
\overline{\mathbf{W}} = \frac{1}{M \ N} \sum_m^M \sum_n^N \mathbf{W}_{m, n}
\end{equation}

\noindent in which each of the elements over the third axis corresponds to the horizontal, vertical and diagonal coefficients of the wavelet transform, respectively. Let furthermore the sum of squared coefficients ${{\Sigma}^2}_{\mathcal{W}_{\psi}}(\mathbf{I})$ be

\begin{equation}\label{eq:sum_of_squares}
{{\Sigma}^2}_{\mathcal{W}_{\psi}}(\mathbf{I}) = \sum_k^3 \sqrt{ \sum_m^M \sum_n^N { \left\{ \mathcal{W}_{\psi}(\mathbf{I}) \right\}_{k, m, n} }^2 }
\end{equation}

For the purposes of this paper, two Daubechies wavelets $\psi_{D7}$ and $\psi_{D8}$, the biorthogonal wavelet $\psi_{B1.5}$ and the Haar wavelet $\psi_{H}$ were used. To prevent spurious signals arising from JPEG compressed source images, a mask was applied onto the images based on the shape of the raw ophthalmoscopic images to filter out any values in the dark areas.

% subsection wavelet_coefficients (end)

% section methodology (end)

\section{Results} % (fold)
\label{sec:results}

In order to understand to what extent image blurriness prejudices diagnostic suitability, the reference dataset by Köhler et al. was used.\cite{kohler2013automatic} This diagnostic data set contains 18 pairs of images, each acquired by a Canon CR-1 fundus camera (Canon Inc. Medical Equipment Group, Tokyo, Japan) with a $45^{\circ}$ field of view, with each pair containing one 'good' and one 'bad' quality image. Each image is stored as a \texttt{3456x5184px} 3-channel RGB JPEG image. Image processing was performed using Python 3.6, primarily using \texttt{skimage} and \texttt{scipy}.\cite{jones2014scipy} The \texttt{pandas} package was used to aggregate and analyse data.\cite{mckinney2011pandas} Wavelets were implemented using the \texttt{PyWavelets} package.\cite{lee2006pywavelets}

\subsection{Principal components analysis (PCA)} % (fold)
\label{sub:principal_components_analysis_pca}

\begin{wrapfigure}[18]{l}{75mm}
% the number in [] of wrapfigure is optional and gives the number of text lines that should be wrapped around the text. Adjust according to your figures height
\includegraphics[width=75mm]{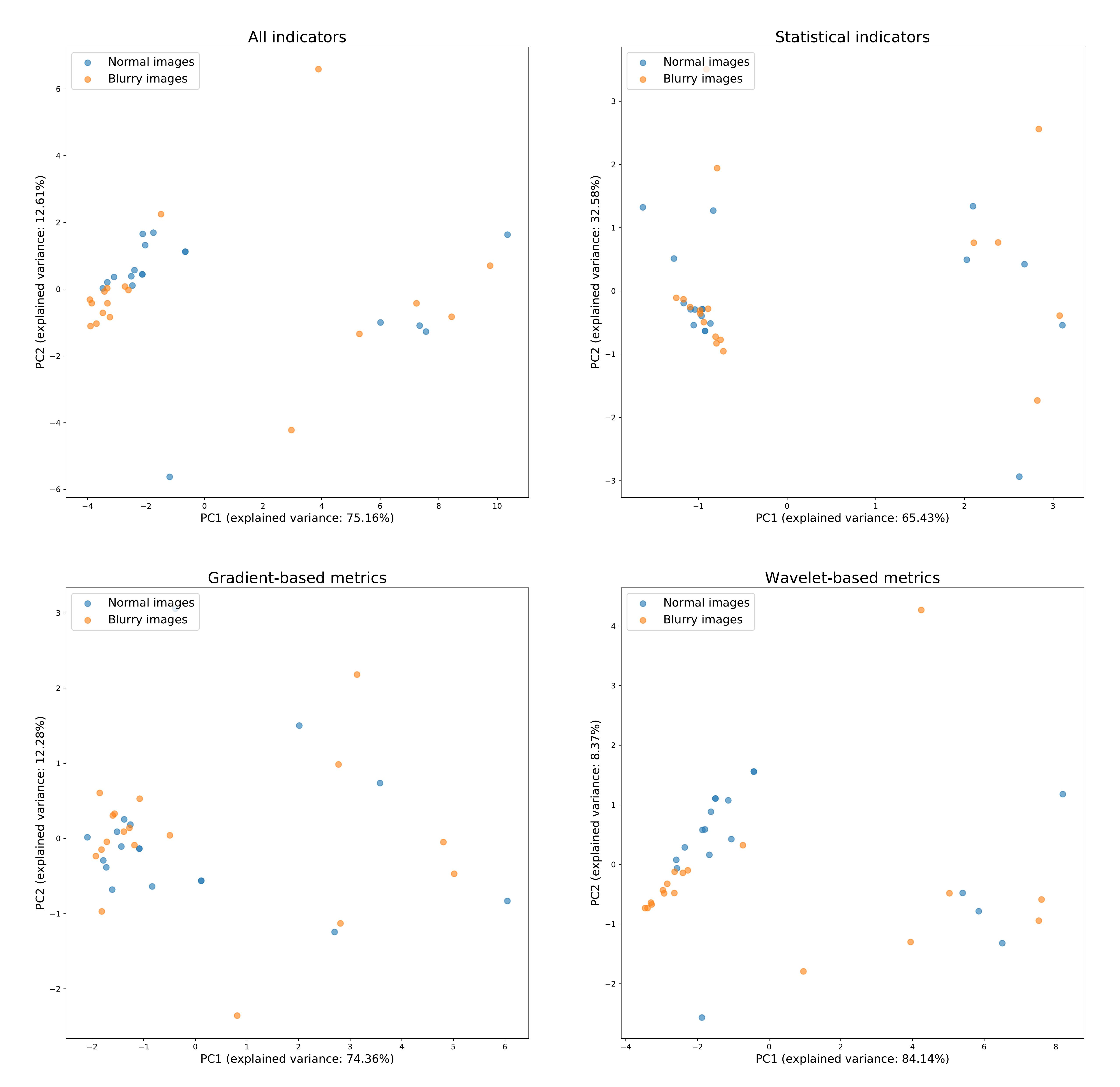}
\captionsetup{labelformat=empty} % makes sure dummy caption is blank
\caption{} % add dummy caption - otherwise \label won't work and figure numbering will not count up
\label{fig3} 
\end{wrapfigure}

% avoid blank space here 
\marginpar{
\vspace{.7cm} % adjust vertical position relative to text with \vspace{} - note that you can enter negative numbers to move margin caption up
\color{Gray} % this gives caption a grey color to set it apart from text body
\textbf{Figure \ref{fig3}. 2-dimensional principal components analysis of all feature variables.} % note that \ref{fig1} refers to the corresponding wrapfigure
 Note the relatively tight groupings of normal-valued images' PCAs surrounded by blurry images' PCAs. 
}
 % avoid blank space here
A PCA was performed to elicit grouping density and dimensional reducibility. PCA was conducted on standard-scaled values, where the standard score $z(X_i)$ of  each value $x$ from a sample $X_{0 \cdots n \mid i \leq n}$ was defined as

\begin{equation}\label{eq:standard_scaler}
	z(x) = \frac{1}{\sigma} \left( x_i - \frac{1}{n} \sum_{k = 1}^n x_k \right)
\end{equation}

The principal components were separately analysed for three distinct variable clusters: statistical indicators (Atkinson's EFC and NEFC metrics, $\overline{E}(\mathbf{I})$ and Shannon entropy $S(\mathbf{I})$), gradients and gradient-derived indicators (Tenengrad, $\Vert\nabla^2\Vert$, perivascular weighted Tenengrad and perivascular weighted $\Vert\nabla^2\Vert$) and for wavelet transforms, the variance of the diagonal, vertical and horizontal coefficients and the sum of squared coefficients by wavelet (\autoref{fig3}).
% subsection principal_components_analysis_pca (end)

\subsection{Machine learning algorithms over the image metrics} % (fold)
\label{sub:machine_learning_algorithms_over_the_image_metrics}

\begin{wrapfigure}[18]{l}{75mm}
\includegraphics[width=75mm]{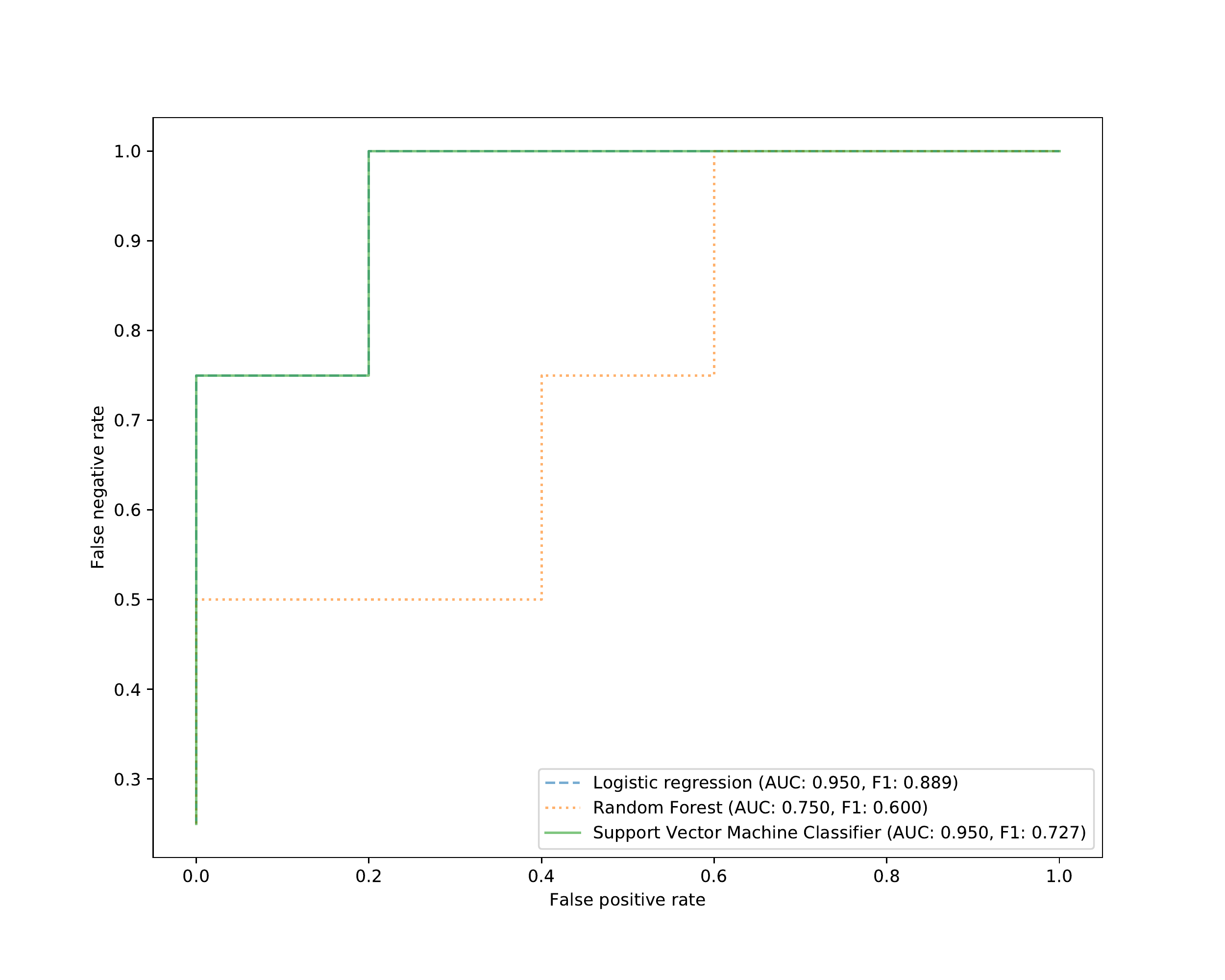}
\captionsetup{labelformat=empty} % makes sure dummy caption is blank
\caption{} % add dummy caption - otherwise \label won't work and figure numbering will not count up
\label{fig:images_all_ROC} 
\end{wrapfigure}

The data set was randomly split into a 25\% ($n=9$) test set and a 75\% ($n=27$) training set. It is important to note that the scarcity of data was suboptimal and for a more thorough understanding of the predictive potential of each of the indicators, as would be discernible from e.g. Bayesian logistic regression using a Monte Carlo sampling algorithm like NUTS (No U-Turn Sampling)\cite{xu2014distributed,hoffman2014no} or simple Metropolis-Hastings sampling, a larger dataset would be indispensable.
% avoid blank space here 
\marginpar{
\vspace{.7cm} % adjust vertical position relative to text with \vspace{} - note that you can enter negative numbers to move margin caption up
\color{Gray} % this gives caption a grey color to set it apart from text body
\textbf{Figure \ref{fig:images_all_ROC}. Comparative performance of classifiers as exemplified by ROC curves.} % note that \ref{fig1} refers to the corresponding wrapfigure
 Note the outstanding performance of both the sigmoid kernel SVM and the crossvalidated logistic regression. Both the sigmoid kernel SVM classifier and the crossvalidated logistic regression arrived at the same boundaries and thus overlap on the ROC curve. This is largely due to the scarce test set.
}
 % avoid blank space here

In this stage, three classifiers were applied:

\begin{itemize}
	\item logistic regression with 5-fold crossvalidation and F1 optimised scoring,
	\item a random forest classifier with 100 estimators, and
	\item a support vector classifier with a sigmoid kernel.
\end{itemize}	

Overall, the performance of both the sigmoid kernel SVM classifier and logistic regression has been excellent, with an Area Under Curve (AUC) of 0.95 for both. However, when considering the $F_1$ score, a metric that considers both accuracy and recall, logistic regression ($F_1 = 0.8889$) far outperforms the sigmoid kernel SVM classifier ($F_1 = 0.7273$) and leaves the random forest model ($F_1 = 0.6000$) far behind. At least partially at fault for this is the relatively small sample count, and it would no doubt be intriguing to repeat these assessments with a larger data set of professionally annotated blurry versus sharp images. 

At the same time, the conclusion is inescapable: by the combination of simple statistical indicators, together with vascularity weighted gradients and wavelet-based metrics drawn from an ensemble of relatively divergent wavelets ($\psi_{D7}$, $\psi_{D8}$ and $\psi_H$ are asymmetric orthogonal bi-orthogonal wavelets while $\psi_{B1.5}$ is a symmetric non-orthogonal biorthogonal wavelet), a fast and efficient method for assessing image blurriness can be generated that is able to triage ophthalmoscopic images with great certainty. For the future of teleophthalmoscopy, the development, standardisation and testing of such an ensemble algorithm will no doubt be a significant first step.

% subsection machine_learning_algorithms_over_the_image_metrics (end)

% section results (end)

\section{Discussion} % (fold)
\label{sec:discussion}

Already a decade ago, Hossain et al. labelled obesity and diabetes in the developing world "a growing challenge".\cite{hossain2009obesity} The emergence of this growing public health issue, fuelled by dietary and cultural changes especially in urban areas in developing nations, will bring with itself a new and unprecedented challenge: ensuring that populations already underserved when it comes to ophthalmic care will receive adequate care for diabetic retinopathy, the prevalence of which can only be expected to increase.\cite{taylor2001world,friedman2011diabetic} Teleophthalmology can be a cornerstone of providing this care to a wide range of the population, even in areas where specialist ophthalmologic resources are scarce, unavailable on site, or difficult to access.\cite{pasquel2016cost,bashshur2015empirical} While smartphone ophthalmoscopy holds enormous potential to provide screening services to underserved populations, its potential beneficial impact is contingent on the ability to acquire high-quality images. Especially where images are acquired with nonspecialist equipment, such as smartphones, pre-analysis Quality Assurance, performed at point of image acquisition, is going to be the crucial determinative factor of whether a healthcare system can successfully leverage teleophthalmology to alleviate the increasing workload arising from the rise in diabetic retinopathy.\cite{mohammadpour2017smartphones,kumar2012teleophthalmology,ye2014global} As computing power in smartphones increases rapidly while their cost rapidly decreases (along with the decreasing cost and increasing ease of access to smartphone ophthalmoscopic adapters, some of which can now be manufactured using a commercial off-the-shelf 3D printer\cite{myung20143d}), the case for on-device quality control becomes increasingly warranted. Since the operational concept of smartphone ophthalmoscopy envisages such devices being used by nonspecialists, such as general practitioners, nurses, health visitors and other public health workers, device-based feedback will be the primary method of quality control and quality triage to ensure that only images that are of sufficient quality to be further evaluated by a specialist are further transmitted into the teleophthalmology system. Fast and high-performance detection of blurry, unsharp or out of focus images is a crucial first step towards ensuring this.

This paper disclosed an ensemble method of multi-component feature extraction, followed by training and evaluating the results on some of the most frequently used machine learning algorithms. Unfortunately, while the initial results are encouraging, the small sample size (a balanced set of $n=36$) draws attention to the fact that more data is needed to isolate the most influential components that assist in determining whether an image is too blurry to be diagnostically useful. It is also important to note that the images used in this project were nonpathological, i.e. taken of generally healthy persons. Testing whether the algorithms are robust with respect to pathologies, including those that manifest as large, diffuse and undifferentiated areas (such as the large white reflex spots typical of retinoblastoma or the diffuse appearance of the optic disc in severe acute papilloedema) is paramount before they can play a valuable clinical role. Given the magnitude of the challenge inherent in providing quality ophthalmic care to a growing population in need and already underserved in many respects, creating a large, professionally annotated data set for the detection of various image quality defects should be considered a global research priority. 

% section discussion (end)

\section*{Acknowledgments}

The author wishes to thank Tam\'{a}s Marton and Katie von Csefalvay for their helpful suggestions in finalising the manuscript, and his colleagues at Starschema for fostering his interest in the subject and encouraging the work that led to this article. All errors and omissions are the author's own.

\subsection*{Competing interests} % (fold)
\label{sub:competing_interests}

The author has no relevant affiliations or financial involvement with any organization or entity with financial interest or financial conflict with the subject matter or materials discussed in the manuscript apart from those disclosed.

% subsection competing_interests (end)

\bibliography{gng-retinopathy}

\bibliographystyle{abbrv}

\end{document}